\pdfoutput=1

\documentclass[11pt,a4paper]{article}
\usepackage[hyperref]{acl2023}
\usepackage{times}
\usepackage{latexsym}
\usepackage{booktabs}
\usepackage{multicol}
\usepackage{multirow}
\usepackage{graphicx}
\usepackage{tikz}
\usepackage[T1]{fontenc}

\usepackage[utf8]{inputenc}

\usepackage{microtype}

\usepackage{inconsolata}

\title{An Annotated Dataset for Explainable Interpersonal Risk Factors of Mental Disturbance in Social Media Posts}

\author{$^{*}$Muskan Garg, $^{**}$Amirmohammad Shahbandegan, $^{\dagger}$Amrit Chadha, $^{**}$Vijay Mago\\
$^{*}$Mayo Clinic, Rochester, MN 55901, USA \\
  $^{**}$Lakehead University,
    Thunder Bay, ON P7B 5E1, Canada \\
  $^{\dagger}$Thapar Institute of Engineering \& Technology,
  Patiala, PB 147005, India}

\begin{document}
\maketitle
\begin{abstract}
With a surge in identifying suicidal risk and its severity in social media posts, we argue that a more consequential and \textit{explainable} research is required for optimal impact on clinical psychology practice and personalized mental healthcare. The success of computational intelligence techniques for inferring mental illness from social media resources, points to natural language processing as a \textit{lens} for determining Interpersonal Risk Factors (IRF) in human writings. Motivated with limited availability of datasets for social NLP research community, we construct and release a new annotated dataset with human-labelled explanations and classification of \textit{IRF} affecting mental disturbance on social media: (i) Thwarted Belongingness (\textsc{TBe}), and (ii) Perceived Burdensomeness (\textsc{PBu}). We establish baseline models on our dataset facilitating future research directions to develop real-time personalized AI models by detecting patterns of \textsc{TBe} and \textsc{PBu} in emotional spectrum of user's historical social media profile. 
\end{abstract}

\section{Introduction}

\begin{figure}
    \centering
    \includegraphics[width=0.90\columnwidth]{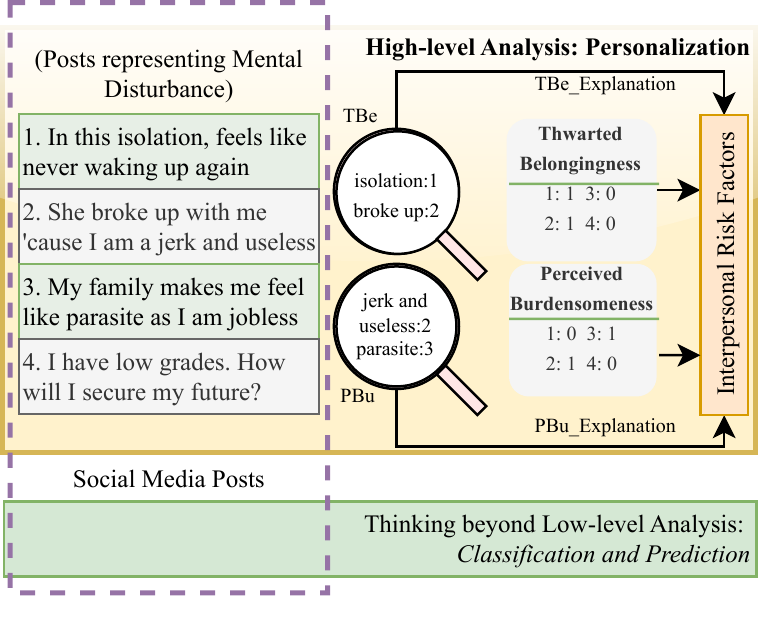}
    \caption{Overview of the problem formulation depicting the need of identifying interpersonal risk factor in texts. The texts [1-4] are annotated as 0: absence or 1: presence of the interpersonal risk factors TBe and PBu.}
    \label{fig:overview}
\end{figure}

The World Health Organization (WHO) emphasizes the importance of significantly accelerating suicide prevention efforts to fulfill the United Nations’ Sustainable Development Goal (SDG) objective by 2030~\cite{saxena2021countdown}.
Reports released in August 2021\footnote{https://www.theguardian.com/society/2021/aug/29/strain-on-mental-health-care-leaves-8m-people-without-help-say-nhs-leaders} indicate that 1.6 million people in England were on waiting lists for mental health care. An estimated 8 million people were unable to obtain assistance from a specialist, as they were not considered \textit{sick enough} to qualify. As suicide remains one of the leading causes of the death worldwide\footnote{https://news.un.org/en/story/2021/06/1094212}, this situation underscores the need of mental health interpretations from social media data where people express themselves and their thoughts, beliefs/emotions with ease~\cite{wongkoblap2022social}. 
The individuals dying by suicide hinder the psychological assessments where a self-reported text or personal writings might be a valuable asset in attempting to assess an individual’s specific personality status and mind rationale~\cite{garg2023mental}. With strong motivation of thinking beyond low-level analysis, Figure~\ref{fig:overview} suggests \textit{personalization} through higher-level analysis of human writings. As, the social media platforms are frequently relied upon as open fora for honest disclosure~\cite{resnik2021naturally}, we examine mental disturbance in Reddit posts aiming to discover Interpersonal Risk Factors (IRF) in text.

\begin{table*}[]
\small
    \centering
    \begin{tabular}{p{3.5cm}p{2cm}|p{1.5cm}p{0.5cm}|p{6cm}|p{0.7cm}}
         \toprule[1.5pt]
        \textbf{Dataset} &  \textbf{Media} & \textbf{Size} & \textbf{Exp.}  & \textbf{Task} & \textbf{Avail.}\\
        \midrule
        \cite{kivran2014understanding} & Twitter & $4454$  & $\times$  & Responses to expressions of loneliness & No \\ 
        \cite{badal2021words} & Interviews & $97$ adults  & $\times$  & Isolation and loneliness in older adults & No \\
        \cite{mahoney2019feeling} & Twitter & $22477$ & $\times$ & Loneliness disclosures throughout the day & No \\
        \cite{ijcai2022p704} & Suicide Notes & $350$ notes & $\times$ & \textsc{TBe} and \textsc{Pbu} in Suicide Notes & OR\\
        \midrule
        Ours & Reddit & $3522$ & \textsc{Yes} & Explainable \textsc{TBe} and \textsc{Pbu} in Social Media Posts & \textsc{Yes}\\
          \bottomrule[1.5pt]
    \end{tabular}
    \caption{Historical evolution of language resources for classifying lonesomeness in texts. OR: On Request, \textsc{TBe}: Thwarted Belongingness and \textsc{Pbu}: Perceived Burdensomeness }
    \label{tab:1}
\end{table*}

Interpersonal relationships are the strong connections that a person with their closest social circle (peers, intimate-partners and family members) which can shape an individual's behavior and range of experience~\cite{puzia2014early}. Affecting such interpersonal relationships influences the associated risk factors resulting in mental disturbance. According to \emph{interpersonal-psychological theory of suicidal behavior}~\cite{joiner2005people}, suicidal desire arises when a person experience persistent emotions of (i) Thwarted Belongingness (\textsc{TBe})\footnote{An unpleasant emotional response to distinguished isolation through mind and character}
, and (ii) Perceived Burdensomeness (\textsc{PBu})\footnote{Characterized by apperceptions that others would 'be better off if I were gone,' underlying unwelcoming society}. As a starting point for our research, this cross-sectional study facilitates the language resource for discovery of underlying users with prospective self-harm/suicidal tendencies to support and compliment existing literature~\cite{bialer2022detecting,tsakalidis2022overview,gaur2018let} as intrinsic classification task.


Computational approaches may better understand the technological advancements in psychology research, aiding in the early detection, prediction and evaluation, management and follow-up of those experiencing suicidal thoughts and behaviors. Most automated systems require available datasets for computational advancements. Past studies show that the availability of relevant datasets in mental healthcare domain is scarce for IRF due to sensitive nature of data as shown in Table~\ref{tab:1}~\cite{su2020deep,garg2023mental}. To this end, we introduce an annotated Reddit dataset for classifying \textsc{TBe} and \textsc{PBu}. The explanatory power of this dataset lies in supporting the motivational interviewing and mental health triaging where early detection of potential risk may trigger an alarm for the need of a mental health practitioner.
We adhere to ethical considerations for constructing and releasing our dataset publicly on Github\footnote{\url{https://github.com/drmuskangarg/Irf}}.


\section{Dataset}


\subsection{Corpus Construction}
\citet{haque2021deep}
used two subreddits $r/depression$ and $r/suicidewatch$ to scrape the SDCNL data and to validate a label correction methodology 
through manual annotation of this dataset for \textit{depression} versus \textit{suicide}. They addressed the then existing ethical issues impacting
\textit{dataset availability} with public release of their dataset. In addition to $1896$ posts of SDCNL dataset, we collected $3362$ additional instances from Reddit on $r/depression$ and $r/SuicideWatch$ through PRAW API\footnote{https://praw.readthedocs.io/en/stable/} from $02$ December $2021$ to $04$ January $2022$ with about $100$ data points per day (to maintain variation in the dataset). On initial screening, we found (i) posts with no self-advocacy, (ii) empty/irrelevant posts. We manually filter them to deduce self-advocacy in texts leveraging $3155$ additional samples, which results in a total of $5051$ data points~\cite{garg2022cams}.
We removed $694$ of the data points depicting no assessment of mental disturbance. Moreover, people write prolonged texts when they indicate IRF which is inline with the conventional arguments where prolonged remarks get better responses from others in comparison of the transient remarks~\cite{park2015manifestation}. The length of real-time Reddit posts varies from a few characters to thousands of words. We limit the maximum length of every post to $300$ words resulting in $3522$ posts as a final corpus.


\subsection{Annotation Scheme}
Classification of IRF, being a complex and highly subjective task, may induce errors with naive judgment. To mitigate this problem, we build a team of three experts: (i) \textit{a clinical psychologist} for training annotators and validating annotations with psychological viewpoint, (ii) \textit{a rehabilitation counselor} for comprehending human mind to understand users' IRF, and (iii) \textit{a social NLP expert} suggesting text based markings in Reddit posts. To negotiate and mitigate the trade-off between three different perspectives, our experts build annotation guidelines\footnote{Please see the Annotation Guidelines in Appendix~\ref{annotations}.} to mark (i) \textsc{TBe}, and (ii) \textsc{PBu}. The experts annotated $40$ samples of the corpus in isolation using these annotation guidelines to avoid biases and discover possible dilemmas due to the subjective nature of tasks. Therefore, we accommodate perplexity guidelines to simplify the task and facilitate unbiased future annotations.
\begin{enumerate}
    \item \textbf{\textsc{TBe} or \textsc{PBu} in the Past}: To check if the condition of a person with disconnected past is still alarming prospect of self-harm or suicidal risk. For instance, `\textit{I was so upset being lonely before Christmas and today I am celebrating New Year with friends'}. We frame rules to handle risk indicators about the past because a person attends celebration and overcome the preceding mental disturbance which means filling void with external event. With neutral opinion by NLP expert about double negation, our clinical psychologist argues presence of risk in their perception which may again evolve after some time and thus, marks this post with presence of the TBe.
    \item\textbf{Ambiguity with \textit{Social Experiences}}: Relationships point to the importance of the ability to take a societal pulse on a regular basis, especially in these unprecedented times of pandemic-induced distancing and shut-downs. People mention major societal events such as breakups, marriage, best friend related issues in various contexts suggesting different user perceptions. We mitigate this problem with two statements: (i) Any feeling of void/missing/regrets/or even mentioning such events with negative words should be marked as presence of TBe such as consider this post: \textit{`But I just miss her SO. much. It's like she set the bar so high that all I can do is just stare at it.'}, (ii) Anything associated with fights/quarrels/general stories should be marked with absence of TBe such as consider the post: \textit{`My husband and I just had a huge argument and he stormed out. I should be crying or stopping him or something. But I decided to take a handful of benzos instead.'}
    
\end{enumerate}

\subsection{Annotation Task}
Three postgraduate students underwent eight hours of professional training by a senior clinical psychologist leveraging annotation and perplexity guidelines. After three successive trial sessions to annotate 40 samples in each round, we ensured their alignment on interpreting task requirements and deployed them for annotating all data points in the corpus. We obtain final annotations based on the majority voting mechanism for binary classification task <\textsc{TBe}, \textsc{PBu}>.\footnote{Sample of dataset is given in Appendix~\ref{sample}.} We validate three annotated files using Fliess' Kappa inter-observer agreement study on classifying \textsc{TBe} and \textsc{PBu} where kappa is calculated as $78.83\%$ and $82.39\%$, respectively.  

Furthermore, we carry out an inter-annotator agreement study with group annotations\footnote{A group of three student annotators extracting explanations and generating a final lists of explanations for \textsc{TBe} as <\textsc{TBe\_exp}> and for \textsc{PBu} as <\textsc{PBu\_exp}>} for text-spans extraction in positive data points. The results for agreement study in two-fold manner: (i) 2 categories (agree, disagree) and (ii) 4 categories (strongly agree, weakly agree, weakly disagree, strongly disagree), are obtained as 82.2\% and 76.4\% for agreement study of <\textsc{TBe\_exp}>, and 89.3\% and 81.3\% for agreement study of <\textsc{PBu\_exp}>, respectively.


\subsection{Dataset Statistics}
 
\begin{table}[]
\small
    \centering
    \begin{tabular}{lll}
         \toprule[1.5pt]
        \textsc{Criteria}& \textsc{Absent} & \textsc{Present}\\
         \toprule[1pt]
        \textsc{Thwarted Belongingness}\\
        \midrule
        Number of Posts & 1595 & 1927\\
        Avg. \#(Words) & 134.68 & 132.58\\
        Avg. \#(Sentences) & 7.73 & 7.61\\
        Max. number of Sentences & 49 & 49\\
        Avg. \#(Words) in Explanations & -  & 3.45\\
         \bottomrule[1.5pt]
        \textsc{Perceived Burdensomeness}\\
        \midrule
        Number of Posts & 2375 & 1147\\
        Avg. \#(Words) & 132.98 & 136.54\\
        Avg. \#(Sentences) & 7.65 & 7.79\\
        Max. number of Sentences & 49 & 32\\
        Avg. \#(Words) in Explanations & -  & 4.04\\
      
      \bottomrule[1.5pt]
    \end{tabular}
    \caption{The statistics of Reddit dataset to determine presence or absence of \textsc{TBe} and \textsc{PBu} and its explanation.}
    \label{tab:2}
\end{table}

\begin{table*}[t!]
\small
\centering
\caption{Comparison of SOTA baseline models' performance}
\label{resultss}
\begin{tabular}{l|ccc|c|ccc|c}
 \toprule[1.5pt]
\textbf{Model}      & \multicolumn{4}{c}{\textsc{Thwarted Belongingness}}& \multicolumn{4}{c}{\textsc{Perceived Burdensomeness}} \\
& Precision & Recall & F1-score & Accuracy& Precision & Recall & F1-score & Accuracy  \\ \midrule[1pt]
\textbf{LSTM}       &  61.40 & 92.77 & 72.00 & 63.67 & 44.65 & 80.90     & 54.69      & 62.35       \\
\textbf{GRU}     &  63.57    & 91.26 & \textbf{73.06} & \textbf{66.70} & 60.87 & 74.77     &  \textbf{63.75}      &    \textbf{78.90 }    \\
\midrule
\textbf{BERT}    & 69.70  &  76.97    & 72.30 & 68.97 & 56.47     &  53.00 &   52.20   &    72.56     \\
\textbf{RoBERTa}    & 71.23 & 73.54   & 71.35   & 68.97 & 67.27       & 37.52       &   45.51 &  74.93     \\
\textbf{DistilBERT}       & 70.24 & 74.08 & 71.15 & 68.50 & 51.15 & 31.89 & 36.93 &      71.71  \\
\textbf{MentalBERT}    & 77.97 & 77.40 & \textbf{76.73} & \textbf{75.12}   &   64.22    &  65.75      & \textbf{62.77}   & \textbf{78.33}    \\
\midrule
\textbf{OpenAI+LR}  & 79.00 &	83.59 &	\textbf{81.23} &	78.62 &	82.66 &	63.08 &	71.55 &	84.58     \\
\textbf{OpenAI+RF}  & 79.06 &	80.68 &	79.86 &	77.48 &	83.33 &	49.23 &	61.90 &	81.36    \\
\textbf{OpenAI+SVM} & 81.31 &	80.34 &	80.83 &	\textbf{78.90} &	79.15 &	74.77 &	\textbf{76.90} &	\textbf{86.19}    \\
\textbf{OpenAI+MLP} & 81.40 &	75.56 &	78.37 &	76.92 &	72.08 &	77.85 &	74.85 &	83.92    \\
\textbf{OpenAI+XGB} & 81.22 &	79.83 &	80.52 &	78.62 &	80.36 &	68.00 &	73.67 &	85.05    \\

 \bottomrule[1.5pt]
\end{tabular}
\end{table*}

\begin{table}[]
    \centering
    \small
    \begin{tabular}{c|c|ccc}
        \toprule[1.5pt]
         Model & Task & P & R & F1 \\
         \midrule 
         LIME & \textsc{TBe} & 14.24  & 53.05 & 20.88 \\
         & \textsc{PBu} & 18.47 & 46.83 & 25.18 \\
         \midrule
         SHAP & \textsc{TBe} & 15.74 & 50.16 & 22.27 \\
         & \textsc{PBu} & 20.77 & 49.89 & 27.92 \\ 
          \bottomrule[1.5pt]
    \end{tabular}
    \caption{Performance Evaluation of explanations of MentalBERT model through LIME and SHAP.}
    \label{tab:5}
\end{table}
 On observing the statistics of our dataset in Table \ref{tab:2}, we found 54.71\% and 32.56\% of positive data points with underlying 255489 and 156620 words for \textsc{TBe} and \textsc{PBu}, respectively.
 It is interesting to note that although the average number of sentences to express \textsc{PBu} is less than \textsc{TBe}, the observations are different for average number of words.
We calculate the Pearson Correlation Coefficient (PCC) for our cross-sectional study on \textsc{TBe} and \textsc{PBu} as 0.0577 which shows slight correlation between the two. Our dataset paves the way for longitudinal studies which is expected to witness increased PCC due to wide spread emotional spectrum~\cite{kolnogorova2021perceived,harrigian2020models}.  
On changing \textsc{TBe} from absence to presence, we observe high rate of increase in positive data points of \textsc{PBu} (((675 - 472)/472) which is 43.00\%) as compared to the absence of \textsc{PBu} (((1252-1123)/1123) which is 11.48\%) suggesting the probability of high correlation in the presence of \textsc{TBe} and \textsc{PBu}, respectively which are given in Table~\ref{tab:7}.
\begin{table}[!ht]
\small
    \centering
    \begin{tabular}{c|cc}
        \midrule \midrule 
       & \textsc{PBu: 0} & \textsc{PBu: 1}\\
        \midrule \midrule
        \textsc{TBe: 0} & 1123 & 472 \\
        \midrule
        
        \textsc{TBe: 1} & 1252 & 675 \\
        \midrule
        \%$\Delta$ & 129/1123 $=0.1148$ & 203/472 $=0.4301$
        \\
         \midrule \midrule 
    \end{tabular}
    \caption{Dataset statistics for Thwarted Belongingness and Perceived Burdensomeness.}
    \label{tab:7}
\end{table}

The most frequent words for identifying (i) \textsc{TBe} are \textit{alone, lonely, nobody to talk, someone, isolated, lost}, and (ii) \textsc{PBu} are \textit{die, suicide, suicidal, kill, burden, cut myself}.\footnote{WordCloud is given in Appendix~\ref{wordcloud}.} Our approach for identifying TBe and PBu goes beyond a simple keyword detector. Instead, we utilize a more sophisticated method that considers the context and relationships between words. For instance, consider a following sample:
\begin{quote}
    Massive party at a friend's house- one of my closest friends is there, loads of my close friends are there, i wasn't invited. wasn't told. only found out on snapchat from their stories. spending new years eve on teamspeak muting my mic every time i break down :)
\end{quote}
Despite the absence of trigger words, our approach flags this post as positive for TBu based on its indicators ‘friend’, ‘teamspeak’, ‘friends’, ‘invited’, ‘snapchat’, to name a few. 



\section{Experiments and Evaluation}

\subsection{Baselines}

We perform extensive analysis to build baselines with three different conventional methods. We first apply \textbf{Recurrent neural networks} where a given text, embedded with GloVe 840B-300\footnote{https://nlp.stanford.edu/projects/glove/}, is sent to a 2-layer RNN model (LSTM, GRU) with 64 hidden neurons and the output is forwarded to two separate fully connected heads: (i) \textsc{TBe} and (ii) \textsc{PBu}. Each of the fully connected blocks have one hidden layer with $16$ neurons and ReLU activation function, and an output layer with sigmoid activation. The loss function is \textit{Binary\_CrossEntropy} and optimizer is \textit{adam} with $lr = 0.001$. Next, we apply \textbf{pretrained transformer-based models}. The input is tokenized using a pre-trained transformers' tokenizer to obtain a 768-dimensional vector which is then fed to a similar fully connected network as the previous architecture with hidden layer size as $48$. We experimented with \textit{roberta-base,
bert-base-uncased,
distilbert-base-uncased, and
mental/mental-bert-base-uncased} models. Finally, we use the \textbf{OpenAI embeddings API}\footnote{https://beta.openai.com/docs/guides/embeddings/embedding-models} to convert the input text into $1536$-dimensional embeddings through `\textit{text-embedding-ada-002}' engine which are used to train a classifier. We test the robustness of this approach over: (i) Logistic Regression, (ii) Random Forest, (iii) Support Vector Machine (iv) Multi Layer Perceptron, and (v) XGBoost.
We further use two explainable methods: (i) \textbf{LIME} and (ii) \textbf{SHAP} on one of the best performing transformer-based models, MentalBERT~\cite{ji2022mentalbert}, to obtain the top keywords~\cite{danilevsky2020survey,zirikly2022explaining}. We compare them with the ground truth ROUGE scores for -- Precision (P), Recall (R), and F1-score (F).

\section{Experimental Settings}
\label{expset}
For consistency, we used the same experimental settings for all models and split the dataset into the train, validation, and test sets. All results are reported on the test set, which makes up 30\% of the whole dataset. We used the grid search optimization technique to optimize the parameters. To tune the number of layers (n), we empirically experimented with the values: learning rate (lr): lr \(\in\) \{0.001, 0.0001, 0.00001\} and optimization (O): O \(\in\) \{`Adam', `Adamax', `AdamW'\}  with a batch-size of {16, 32} were used. We used base version pre-trained language models (LMs) using HuggingFace\footnote{https://huggingface.co/models}, an open-source Python library. We used optimized parameters for each baseline to find precision, recall, F1-score, and Accuracy. Varying lengths of posts are padded to 256 tokens with truncation. Each model was trained for 20 epochs, and the best-performing model based on the average accuracy score was saved. Thus, we set hyperparameter for our experiments as $Optimizer$ = Adam, learning rate = 1e-3, batch size= $16$, and epochs=$20$. 



\subsection{Experimental Results}
Table~\ref{resultss} shows the performance of state-of-the-art methods in terms of precision, recall, F1-score, and accuracy. The current models have moderately low performance in this task, possibly due to a lack of ability to capture contextual information in the text. MentalBERT, a transformer-based language model, initialized with BERT-Base and trained with mental health-related posts collected from Reddit, had the best performance among BERT-based models, with an F1-score of 76.73\% and 62.77\% for \textsc{TBe} and \textsc{PBu}, respectively. This is likely due to the fact that it was trained on the same context as the task, namely health-related posts on Reddit. The combination of OpenAI embeddings and a classifier outperforms RNN and transformer-based models. The highest F1-Score of 81.23\% was achieved by logistic regression for \textsc{TBe}, while the best performing model for \textsc{PBu} was SVM with an F1-score of 76.90\%. We also analyzed the explainability of the model using LIME and SHAP methods of explainable AI for NLP on the best performing transformer model (MentalBERT) for \textsc{TBe} and \textsc{PBu}. We obtain results for all positive data points in the testing dataset and observe high recall of text-spans with reference to the ground truth as shown in Table~\ref{tab:5}. We find the scope of improvement by limiting the superfluous text-spans found in the resulting set of words. The consistency in results suggests the need of contextual/domain-specific knowledge and infusing commonsense to improve explainable classifiers for a given task.

\section{Conclusion and Future Work}

We present a new annotated dataset for discovering interpersonal risk factors through human-annotated extractive explanations in the form of text-spans and binary labels in $3522$ English Reddit posts. In future work, we plan to enhance the dataset with more samples and develop new models tailored explicitly to \textsc{TBe} and \textsc{PBu}. The implications of this work include the potential to improve public health surveillance and other mental healthcare applications that rely on automatically identifying posts in which users describe their mental health issues. We keep the implementation of explainable AI models for multi-task text classification, as an open research direction for Open AI and other newly developed responsible AI models. We pose the discovery of new research frontiers for future, through longitudinal study on users' historical social media profile to examine interpersonal risk factors and potential risk of self-harm or suicidal ideation.  As we focus on Reddit data as a starting point of our study, exploring other forums could be an interesting research direction.

\section*{Acknowledgement}

We express our gratitude to Veena Krishnan, a senior clinical psychologist, and Ruchi Joshi, a rehabilitation counselor, for their unwavering support throughout the project. Additionally, we extend our heartfelt appreciation to Prof. Sunghwan Sohn for his consistent guidance and support. This project was partially supported by NIH R01 AG068007. This project is funded by NSERC Discovery Grant (RGPIN-2017-05377), held by Vijay Mago, Department of Computer Science, Lakehead University, Canada.

\section*{Limitations}

There might be linguistic discrepancies between Reddit users and Twitter users who post about their mental disturbance on social media. Social media users may intentionally post such thoughts to gain attention of other social media users but for simplicity, we assume the social media posts to be credible. Thus, we assume that the social media posts are not misleading. We acknowledge that our work is subjective in nature and thus, interpretation about wellness dimensions in a given post may vary from person to person. 

\section*{Ethical Considerations}

The dataset we use is from Reddit, a forum intended for anonymous posting, users’ IDs are anonymized. In addition, all sample posts shown throughout this work are anonymized, obfuscated, and paraphrased for user privacy and to prevent misuse. Thus, this study does not require ethical approval. Due to the subjective nature of annotation, we expect some biases in our gold-labeled data and the distribution of labels in our dataset. Examples from a wide range of users and groups are collected, as well as clearly defined instructions, in order to address these concerns. Due to high inter-annotator agreement (\(\kappa\) score), we are confident that the annotation instructions are correctly assigned in most of the data points. It is reproducible with the dataset and the source code to reproduce the baseline results which is available on Github. 

To address concerns around potential harms, we believe that the tool should be used by professionals who are trained to handle and interpret the results. We recognize the huge impact of false negatives in practical use of applications such as mental health triaging, and we shall continue working towards improving its accuracy and reducing the likelihood of false negatives. We further acknowledge that our work is empirical in nature and we do not claim to provide any solution for clinical diagnosis at this stage. 

\bibliographystyle{acl_natbib}
\bibliography{anthology,custom}

\begin{thebibliography}{22}
\expandafter\ifx\csname natexlab\endcsname\relax\def\natexlab#1{#1}\fi

\bibitem[{Badal et~al.(2021)Badal, Nebeker, Shinkawa, Yamada, Rentscher, Kim,
  and Lee}]{badal2021words}
Varsha~D Badal, Camille Nebeker, Kaoru Shinkawa, Yasunori Yamada, Kelly~E
  Rentscher, Ho-Cheol Kim, and Ellen~E Lee. 2021.
\newblock Do words matter? detecting social isolation and loneliness in older
  adults using natural language processing.
\newblock \emph{Frontiers in Psychiatry}, 12.

\bibitem[{Bialer et~al.(2022)Bialer, Izmaylov, Segal, Tsur, Levi-Belz, and
  Gal}]{bialer2022detecting}
Amir Bialer, Daniel Izmaylov, Avi Segal, Oren Tsur, Yossi Levi-Belz, and Kobi
  Gal. 2022.
\newblock Detecting suicide risk in online counseling services: A study in a
  low-resource language.
\newblock In \emph{Proceedings of the 29th International Conference on
  Computational Linguistics COLING}, pages 4241--4250.

\bibitem[{Danilevsky et~al.(2020)Danilevsky, Qian, Aharonov, Katsis, Kawas, and
  Sen}]{danilevsky2020survey}
Marina Danilevsky, Kun Qian, Ranit Aharonov, Yannis Katsis, Ban Kawas, and
  Prithviraj Sen. 2020.
\newblock A survey of the state of explainable ai for natural language
  processing.
\newblock In \emph{Proceedings of the 1st Conference of the Asia-Pacific
  Chapter of the Association for Computational Linguistics and the 10th
  International Joint Conference on Natural Language Processing}, pages
  447--459.

\bibitem[{Garg(2023)}]{garg2023mental}
Muskan Garg. 2023.
\newblock Mental health analysis in social media posts: A survey.
\newblock \emph{Archives of Computational Methods in Engineering}, pages 1--24.

\bibitem[{Garg et~al.(2022)Garg, Saxena, Krishnan, Joshi, Saha, Mago, and
  Dorr}]{garg2022cams}
Muskan Garg, Chandni Saxena, Veena Krishnan, Ruchi Joshi, Sriparna Saha, Vijay
  Mago, and Bonnie~J Dorr. 2022.
\newblock Cams: An annotated corpus for causal analysis of mental health issues
  in social media posts.
\newblock In \emph{Language Resources Evaluation Conference (LREC)}.

\bibitem[{Gaur et~al.(2018)Gaur, Kursuncu, Alambo, Sheth, Daniulaityte,
  Thirunarayan, and Pathak}]{gaur2018let}
Manas Gaur, Ugur Kursuncu, Amanuel Alambo, Amit Sheth, Raminta Daniulaityte,
  Krishnaprasad Thirunarayan, and Jyotishman Pathak. 2018.
\newblock " let me tell you about your mental health!" contextualized
  classification of reddit posts to dsm-5 for web-based intervention.
\newblock In \emph{Proceedings of the 27th ACM International Conference on
  Information and Knowledge Management}, pages 753--762.

\bibitem[{Ghosh et~al.(2022)Ghosh, Ekbal, and Bhattacharyya}]{ijcai2022p704}
Soumitra Ghosh, Asif Ekbal, and Pushpak Bhattacharyya. 2022.
\newblock \href {https://doi.org/10.24963/ijcai.2022/704} {Am i no good?
  towards detecting perceived burdensomeness and thwarted belongingness from
  suicide notes}.
\newblock In \emph{Proceedings of the Thirty-First International Joint
  Conference on Artificial Intelligence, {IJCAI-22}}, pages 5073--5079.
  International Joint Conferences on Artificial Intelligence Organization.
\newblock AI for Good.

\bibitem[{Haque et~al.(2021)Haque, Reddi, and Giallanza}]{haque2021deep}
Ayaan Haque, Viraaj Reddi, and Tyler Giallanza. 2021.
\newblock Deep learning for suicide and depression identification with
  unsupervised label correction.
\newblock In \emph{International Conference on Artificial Neural Networks},
  pages 436--447. Springer.

\bibitem[{Harrigian et~al.(2020)Harrigian, Aguirre, and
  Dredze}]{harrigian2020models}
Keith Harrigian, Carlos Aguirre, and Mark Dredze. 2020.
\newblock Do models of mental health based on social media data generalize?
\newblock In \emph{Findings of the association for computational linguistics:
  EMNLP 2020}, pages 3774--3788.

\bibitem[{Ji et~al.(2022)Ji, Zhang, Ansari, Fu, Tiwari, and
  Cambria}]{ji2022mentalbert}
Shaoxiong Ji, Tianlin Zhang, Luna Ansari, Jie Fu, Prayag Tiwari, and Erik
  Cambria. 2022.
\newblock Mentalbert: Publicly available pretrained language models for mental
  healthcare.
\newblock In \emph{Proceedings of the Thirteenth Language Resources and
  Evaluation Conference}, pages 7184--7190. European Language Resources
  Association (ELRA).

\bibitem[{Joiner et~al.(2005)}]{joiner2005people}
Thomas~E Joiner et~al. 2005.
\newblock \emph{Why people die by suicide}.
\newblock Harvard University Press.

\bibitem[{Kivran-Swaine et~al.(2014)Kivran-Swaine, Ting, Brubaker, Teodoro, and
  Naaman}]{kivran2014understanding}
Funda Kivran-Swaine, Jeremy Ting, Jed Brubaker, Rannie Teodoro, and Mor Naaman.
  2014.
\newblock Understanding loneliness in social awareness streams: Expressions and
  responses.
\newblock In \emph{Proceedings of the International AAAI Conference on Web and
  Social Media}, volume~8, pages 256--265.

\bibitem[{Kolnogorova et~al.(2021)Kolnogorova, Allan, Moradi, and
  Stecker}]{kolnogorova2021perceived}
Kateryna Kolnogorova, Nicholas~P Allan, Shahrzad Moradi, and Tracy Stecker.
  2021.
\newblock Perceived burdensomeness, but not thwarted belongingness, mediates
  the impact of ptsd symptom clusters on suicidal ideation modeled
  longitudinally.
\newblock \emph{Journal of Affective Disorders}, 282:133--140.

\bibitem[{Mahoney et~al.(2019)Mahoney, Le~Moignan, Long, Wilson, Barnett,
  Vines, and Lawson}]{mahoney2019feeling}
Jamie Mahoney, Effie Le~Moignan, Kiel Long, Mike Wilson, Julie Barnett, John
  Vines, and Shaun Lawson. 2019.
\newblock Feeling alone among 317 million others: Disclosures of loneliness on
  twitter.
\newblock \emph{Computers in Human Behavior}, 98:20--30.

\bibitem[{Park et~al.(2015)Park, Kim, Lee, Yoo, Jeong, and
  Cha}]{park2015manifestation}
Sungkyu Park, Inyeop Kim, Sang~Won Lee, Jaehyun Yoo, Bumseok Jeong, and
  Meeyoung Cha. 2015.
\newblock Manifestation of depression and loneliness on social networks: a case
  study of young adults on facebook.
\newblock In \emph{Proceedings of the 18th ACM conference on computer supported
  cooperative work \& social computing}, pages 557--570.

\bibitem[{Puzia et~al.(2014)Puzia, Kraines, Liu, and Kleiman}]{puzia2014early}
Megan~E Puzia, Morganne~A Kraines, Richard~T Liu, and Evan~M Kleiman. 2014.
\newblock Early life stressors and suicidal ideation: Mediation by
  interpersonal risk factors.
\newblock \emph{Personality and Individual Differences}, 56:68--72.

\bibitem[{Resnik et~al.(2021)Resnik, Foreman, Kuchuk, Musacchio~Schafer, and
  Pinkham}]{resnik2021naturally}
Philip Resnik, April Foreman, Michelle Kuchuk, Katherine Musacchio~Schafer, and
  Beau Pinkham. 2021.
\newblock Naturally occurring language as a source of evidence in suicide
  prevention.
\newblock \emph{Suicide and Life-Threatening Behavior}, 51(1):88--96.

\bibitem[{Saxena and Kline(2021)}]{saxena2021countdown}
Shekhar Saxena and Sarah Kline. 2021.
\newblock Countdown global mental health 2030: data to drive action and
  accountability.
\newblock \emph{The Lancet Psychiatry}, 8(11):941--942.

\bibitem[{Su et~al.(2020)Su, Xu, Pathak, and Wang}]{su2020deep}
Chang Su, Zhenxing Xu, Jyotishman Pathak, and Fei Wang. 2020.
\newblock Deep learning in mental health outcome research: a scoping review.
\newblock \emph{Translational Psychiatry}, 10(1):1--26.

\bibitem[{Tsakalidis et~al.(2022)Tsakalidis, Chim, Bilal, Zirikly,
  Atzil-Slonim, Nanni, Resnik, Gaur, Roy, Inkster
  et~al.}]{tsakalidis2022overview}
Adam Tsakalidis, Jenny Chim, Iman~Munire Bilal, Ayah Zirikly, Dana
  Atzil-Slonim, Federico Nanni, Philip Resnik, Manas Gaur, Kaushik Roy, Becky
  Inkster, et~al. 2022.
\newblock Overview of the clpsych 2022 shared task: Capturing moments of change
  in longitudinal user posts.

\bibitem[{Wongkoblap et~al.(2022)Wongkoblap, Vadillo, and
  Curcin}]{wongkoblap2022social}
Akkapon Wongkoblap, Miguel~A Vadillo, and Vasa Curcin. 2022.
\newblock Social media big data analysis for mental health research.
\newblock In \emph{Mental Health in a Digital World}, pages 109--143. Elsevier.

\bibitem[{Zirikly and Dredze(2022)}]{zirikly2022explaining}
Ayah Zirikly and Mark Dredze. 2022.
\newblock Explaining models of mental health via clinically grounded auxiliary
  tasks.
\newblock In \emph{Proceedings of the Eighth Workshop on Computational
  Linguistics and Clinical Psychology}, pages 30--39.

\end{thebibliography}

\appendix

\section{Sample Dataset}
\label{sample}
The sample dataset is given in Table~\ref{tab:6}.
\begin{table*}[]
    \centering
    \caption{A sample of dataset to examine interpersonal risk factors and their explanations for mental health problems}
    \begin{tabular}{p{10cm}|p{0.5cm}p{1.5cm}|p{0.5cm}p{1.5cm}}
        \midrule \midrule
      
        \textsc{Text}  & \textsc{TBe}  & \textsc{TBe\_exp} &  \textsc{PBu} &\textsc{PBu\_exp}\\
        \midrule \midrule
        To be rather blunt, I'm single, stuck living with parents and working shitty hours. I don't have any friends, I've never been in a proper, loving relationship and I'm a socially awkward loser. Other people see me as a burden, people hate talking to me, and I'm tired of continuing on with this. It's been 10 years since this mess started, do I not deserve a life worth living? & 1 & Social awkward & 1 & See me as a burden\\
        \midrule
        I have lost around 8 friends over the past two years. They leave without even saying goodbye. It's literally just my personality. I'm a "downer" apparently. I'm scared that I'll be alone forever. Should I change so that someone will like me? & 1 & Alone forever & 0 & - \\
        \midrule
        I'm having thoughts about killing myself to escape all of this. Its the most dumb thing to do but i feel like im running out of choices. We're not financially stable. I'm a student. I should have wore a condom. What should i do. & 0 & - & 1 & killing myself \\
        \midrule
        I only take Lexapro. I was watching some videos on these guy that call themselves "Preppers" and they prep for the end of the world. They say that people on any types of drugs will become unstable and focused on getting their fix or whatever. Is that us? & 0 & - & 0 & -\\
        \midrule
         
         \midrule
    \end{tabular}
    
    \label{tab:6}
\end{table*}

\section{Annotation Guidelines}
\label{annotations}
We follow The Interpersonal Needs
Questionnaire (INQ) in association with our experts to set required guidelines. According to the Baumeister and Leary (1995) theory of the \textit{need to belong}, \textbf{thwarted belongingness} (\textsc{TBe}) is a psychologically-painful mental state that results from inadequacy of connectedness. It contains detailed set of instructions to mark latent feeling of disconnectedness, missing someone, major event such as death, or being ignored/ostracized/alienated, as  \textsc{TBe}.

\begin{quote}
    Marking:\\
    0: No Thwarted Belongingness\\
    1: Thwarted Belongingness present
\end{quote}

\textbf{Perceived burdensomeness} (\textsc{PBu}) is a mental state characterized by making fully conscious perception that others would “be better off if I were gone,” which manifests when the need for social competence. The Self-Determination Theory (Ryan \& Deci, 2000) proposes the association of family discord, unemployment, and functional impairment  with suicide across the lifespan. Detailed set of instructions were given to mark the major feeling of \textit{being a burden on other people and/or society}, as \textsc{PBu}.

\begin{quote}
    Marking:\\
    0: No Perceived Burdensomeness\\
    1: Perceived Burdensomeness present
\end{quote}

\textsc{TBe} and \textsc{PBu} are the most proximal mental states that precede the development of thoughts of suicide—stressful life events, mental disorders, and other risk factors for suicide are relatively more distal in the causal chain of risk factors for suicide. These IRF are posited to be dynamic and amenable to therapeutic change.

\section{Word Frequency in Explanations}
\label{wordcloud}
\begin{figure}[!t]
    \centering
    \includegraphics[width=6cm,height=6cm]{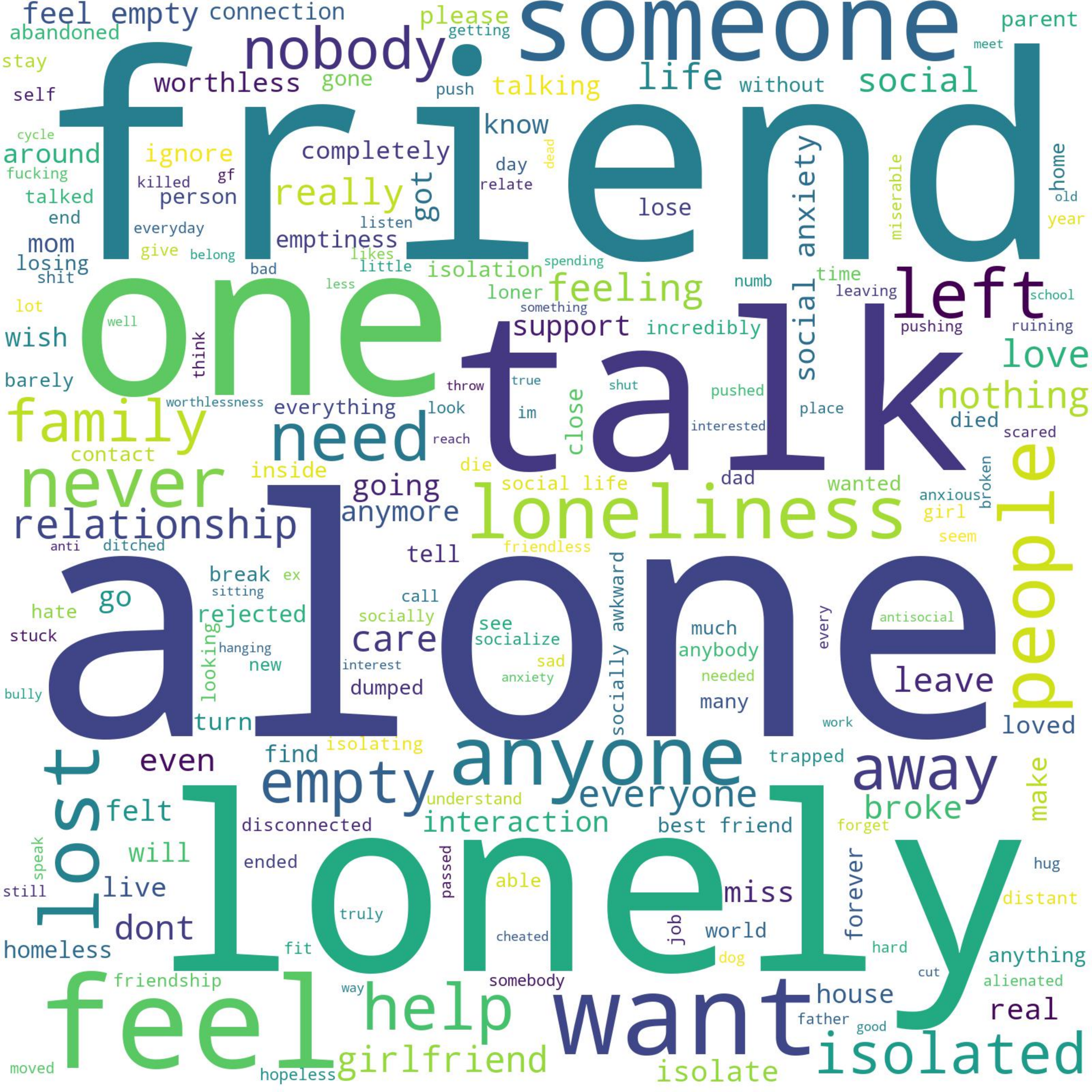}
    \caption{Wordcloud for Thwarted Belongingness}
    \label{fig:3}
\end{figure}

\begin{figure}[!t]
    \centering
    \includegraphics[width=6cm,height=6cm]{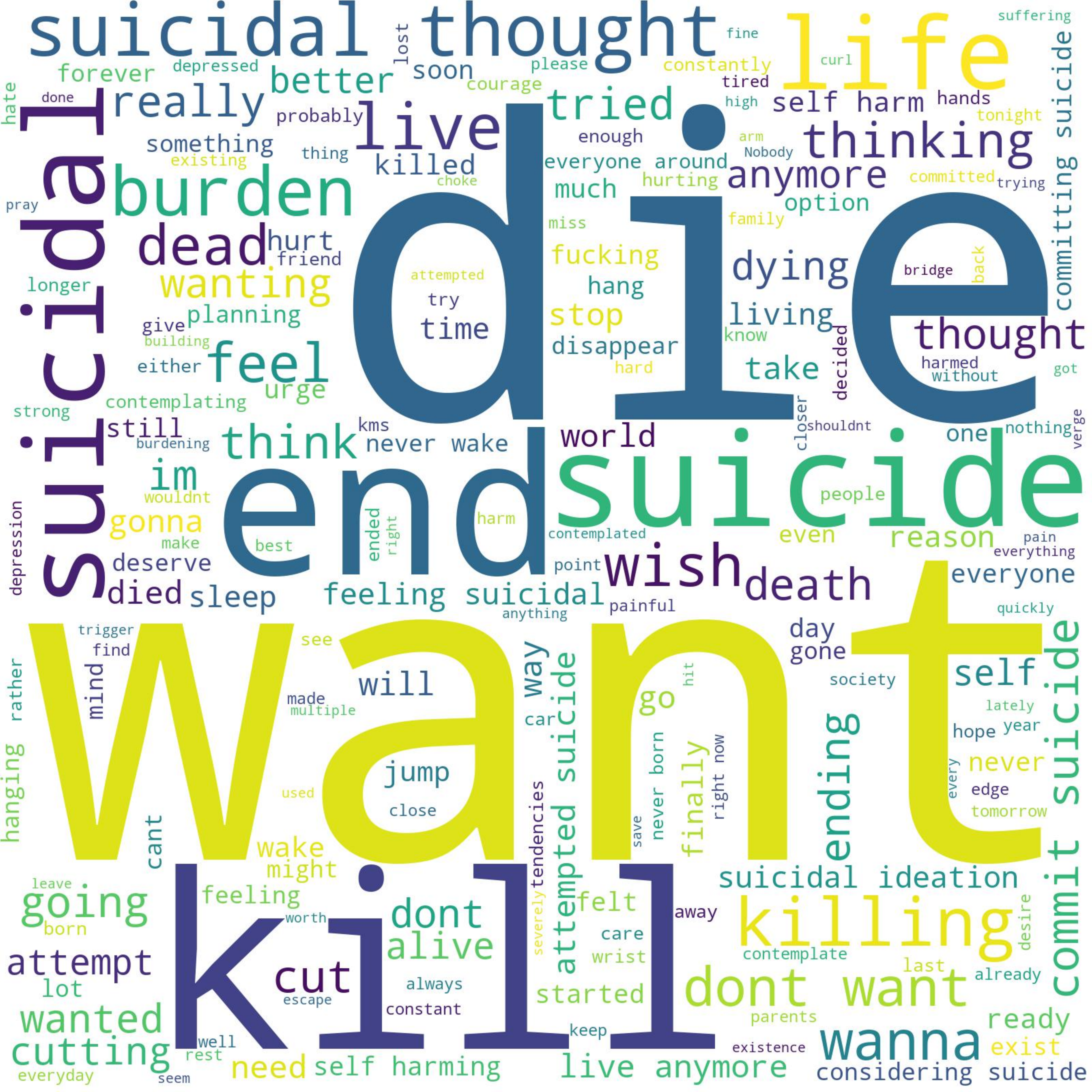}
    \caption{Wordcloud for Perceived Burdensomeness}
    \label{fig:4}
\end{figure}

The wordcloud for explanations are shown in Figures~\ref{fig:3} and \ref{fig:4}.

\end{document}